\documentclass[10pt,twocolumn,letterpaper]{article}

\usepackage[pagenumbers]{cvpr} 

\definecolor{cvprblue}{rgb}{0.21,0.49,0.74}
\usepackage[pagebackref,breaklinks,colorlinks,allcolors=cvprblue]{hyperref}

\usepackage{amsmath}
\usepackage{amssymb}
\usepackage{mathtools}
\usepackage{amsthm}
\usepackage{multirow}
\usepackage[table]{xcolor}
\usepackage[most]{tcolorbox}
\usepackage{microtype}
\usepackage{graphicx}
\usepackage{subcaption}
\usepackage{booktabs}
\usepackage{algorithm}
\usepackage{algorithmic}
\usepackage{listings}
\usepackage{tablefootnote}

\theoremstyle{plain}

\theoremstyle{definition}

\theoremstyle{remark}

\theoremstyle{definition}

\tcolorboxenvironment{observation}{
    breakable,
    colback=gray!10!white,
    boxrule=0pt
}
\tcolorboxenvironment{theorem}{
    breakable,
    colback=gray!10!white,
    boxrule=0pt
}



\title{When AI Designs AI: Innovation or Imitation?}

\author{
  \textbf{Yikang Yang}$^{1,2,4}$ \quad
  \textbf{Zhengxin Yang}$^{1,2,3}$\thanks{Zhengxin Yang is the corresponding author} \quad
  \textbf{Luzhou Peng}$^{2,4}$ \quad
  \textbf{Minghao Luo}$^{5}$ \quad
   \\
  \textbf{Yanqi Kan}$^{2,4}$ \quad
  \textbf{Wanling Gao}$^{2,3}$ \quad
  \textbf{Jianfeng Zhan}$^{2,3}$ \\
  State Key Lab of Processors, Institute of Computing Technology, CAS, Beijing 100190, China$^{1}$ \\
  Institute of Computing Technology, Chinese Academy of Sciences$^{2}$\\
  BenchCouncil (International Open Benchmark Council)$^{3}$\\
  University of Chinese Academy of Sciences$^{4}$\\
  Department of Computer Science McCormick School of Engineering Northwestern University Evanston, IL, USA$^{5}$\\
  \texttt{\{yangyikang23s, yangzhengxin\}@ict.ac.cn}
}

\begin{document}
\maketitle
\begin{abstract}
    Recent advances in LLM agents have made them increasingly capable of designing methods for complex AI tasks.
    This raises two central questions about agent-designed methods relative to human-designed methods: how well they perform, and how different their algorithmic designs are.
    To study these questions, this paper introduces an analysis that derives task-specific algorithmic design spaces from human-designed methods, maps both human- and agent-designed methods into these spaces, and quantifies their algorithmic differences at the module level.
    Widely used LLM agents are evaluated on a suite of representative, open-ended AI tasks spanning multiple modalities, and the methods they design are analyzed in terms of both task performance and algorithmic differences from human-designed methods.
    Experimental results show that current agents can occasionally match or surpass human state-of-the-art (SOTA) performance (10/72 configurations), but such success does not generalize reliably across tasks or agents.
    Moreover, 96.8\% of agent-designed methods fall within human-derived algorithmic design spaces, largely recombining algorithmic choices found in human-designed methods, while nearly half exactly match an existing human algorithmic design.
    Taken together, these findings suggest that although current agents can occasionally match or surpass human SOTA performance, their algorithmic designs remain within human-derived algorithmic design spaces, reflecting the reuse and recombination of algorithmic choices.
\end{abstract}

\section{Introduction}
LLM agents extend chat-only language models into systems capable of using tools and interacting with external environments~\cite{React,Toolformer,MemGPT,Reflexion}.
These advances have enabled LLM agents to tackle increasingly complex tasks across software engineering, web interaction, scientific discovery, and other domains~\cite{SWE-agent, WebArena, AIScientist, MetaGPT,OpenHands}.
These capabilities naturally lead to the idea of using LLM agents to design methods for complex AI tasks.
This raises two central questions about agent-designed methods relative to human-designed methods: \textit{how well they perform, and how different their algorithmic designs are.}

Existing work has developed benchmarks for evaluating LLM agents across a range of AI research and engineering tasks~\cite{MLAgentBench,MLE-Bench,MLGym,AIRS-Bench,RE-Bench}.
Most of these benchmarks primarily assess agents based on final task performance, showing whether they achieve competitive results but offering limited insight into the algorithmic choices behind their performance.
InnoGym is a recent exception that evaluates methodological novelty, but its reliance on LLM-as-a-judge introduces model-dependent bias~\cite{InnoGym}.
In addition, some benchmark tasks offer limited room for open-ended algorithmic exploration, either because their solution paths are constrained or because performance is already close to saturation~\cite{MLAgentBench,MLE-Bench,MLGym,AIRS-Bench}.
These limitations make existing benchmarks insufficient on their own for analyzing the algorithmic choices of agent-designed methods, and their relationship to performance.

To address these limitations, this paper introduces a suite of representative, open-ended AI tasks that span multiple modalities and offer meaningful room for further performance improvement.
Each task includes a human leaderboard with corresponding reference papers and code.
This paper develops a human-in-the-loop workflow for building task-specific algorithmic design spaces from the referenced methods.
Both human and agent-designed methods are then represented as coordinates in the corresponding task-specific design space.
This common representation enables joint analysis of where agents explore, how their methods differ from human designs, and how these choices relate to performance.

\begin{figure*}[ht]
  \centering
  \includegraphics[width=0.95\textwidth]{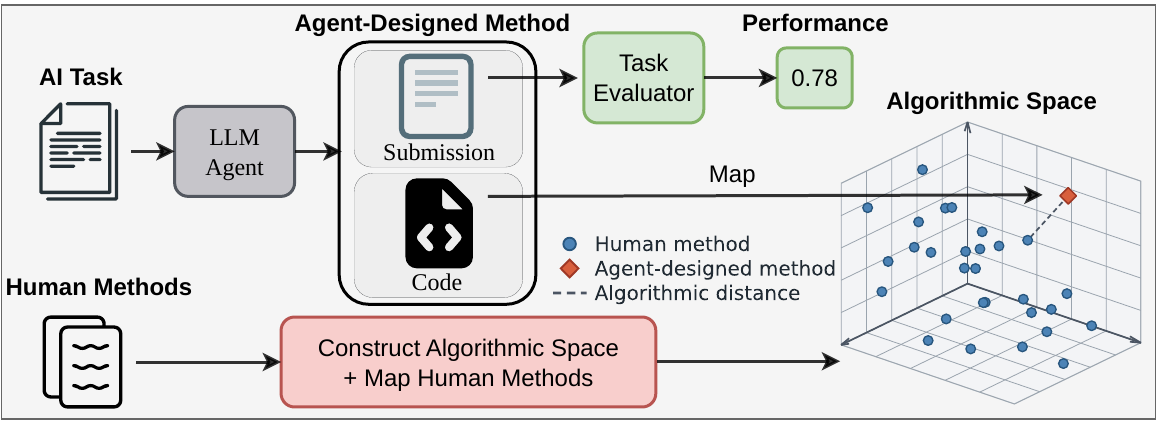}
  \caption{Overview of the analysis methodology. An LLM agent develops and executes code for an AI task, producing submissions that are evaluated by a task-specific evaluator to measure performance. The code is mapped into an algorithmic design space constructed from human reference methods. Human- and agent-designed methods are represented in the same task-specific space, enabling their algorithmic differences to be quantified at the module level.}
  \label{fig:figure1}
\end{figure*}

This study evaluates widely used LLM agents on the task suite.
Current agents can occasionally reach or surpass Human SOTA, but do so in only 10 of the 72 evaluation configurations and on only 3 of the 6 tasks.
Meanwhile, most agent-generated methods can be explained as recombinations or local variations of human methods: nearly half algorithmically match an existing human method, and over 70\% differ from the nearest human method in at most one module.
This study also identifies several recurring patterns in agent research, including limited use of external knowledge, limited exploration of algorithmic design spaces, and a strong preference for ensemble prediction.
These findings suggest that current agents can occasionally reach the human frontier, but their exploration remains centered on established human methods and covers only a limited range of algorithmic designs.

The main contributions of this paper are as follows:
\begin{itemize}
    \item This paper provides a systematic empirical study of how widely used LLM agents design methods for complex AI tasks, analyzing method performance, algorithmic designs, and patterns in agent research.
    \item To support this study, this paper introduces a suite of representative, open-ended AI tasks spanning multiple modalities, together with human leaderboards and corresponding reference papers and code.
    \item This paper develops a human-in-the-loop workflow that derives task-specific algorithmic design spaces from human-designed methods and represents both human- and agent-designed methods within them.
\end{itemize}

The remainder of this paper is organized as follows.
Section~\ref{sec:related-work} reviews related work on LLM agents and benchmarks for AI tasks.
Section~\ref{sec:analysis-methodology} describes the analysis methodology of this paper. 
Section~\ref{sec:experiments} presents the experimental results on agent performance, algorithmic designs, and patterns in agent research.
Section~\ref{sec:discussion} discusses the implications of these findings and possible routes toward more reliable progress.
Section~\ref{sec:conclusion} concludes the paper.
\section{Related Work}
\label{sec:related-work}

\subsection{LLM Agents}

LLM agents extend language models into systems that can observe external environments, invoke tools, and take actions~\cite{LLMAgentSurvey1, LLMAgentSurvey2}.
In early work, ReAct developed a mechanism for LLMs to interact with environments~\cite{React}, while Toolformer focused on enabling language models to use tools~\cite{Toolformer}.
Subsequent work enhanced agents' capabilities in feedback-based self-reflection, long-term memory management, and reusable skill accumulation~\cite{Reflexion,MemGPT,Voyager}.
Today, Claude Code~\cite{ClaudeCode}, Codex~\cite{Codex}, and Gemini CLI~\cite{GeminiCLI} are among the most widely used coding agents.
Academic work has also explored specialized systems for AI research, including AIDE~\cite{AIDE} and MLEvolve~\cite{MLEvolve}, with related efforts using LLMs to iteratively search for and improve candidate programs~\cite{FunSearch,AlphaEvolve}.
Rather than proposing a new agent system, this paper evaluates widely used agents on AI tasks and analyzes the performance and algorithmic designs of the methods they produce.

\subsection{Benchmarks for LLM Agents on AI Tasks}

Existing benchmarks evaluate LLM agents on machine learning experimentation and engineering, AI research challenges, paper replication, and frontier AI R\&D~\cite{MLAgentBench,MLE-Bench,MLGym,MLRC-Bench,PaperBench,AIRS-Bench,RE-Bench,RExBench}.
Most assess agents primarily through task completion and final performance, offering limited insight into the algorithmic choices behind the resulting methods.
Some also include tasks whose solution paths are constrained or whose performance is close to saturation, limiting open-ended algorithmic exploration~\cite{MLAgentBench,MLE-Bench,MLGym}.
InnoGym is a recent exception that evaluates both performance gain and methodological novelty~\cite{InnoGym}.
Its novelty evaluation, however, relies on LLM-as-a-judge comparisons, which can introduce model-dependent bias and inconsistency.
Rather than evaluating novelty, this paper maps human and agent-designed methods into explicit task-specific algorithmic design spaces and measures the coordinate distances between them, making method differences traceable and auditable while partially mitigating subjectivity.
\begin{figure*}[t]
  \centering
  \includegraphics[width=0.95\textwidth]{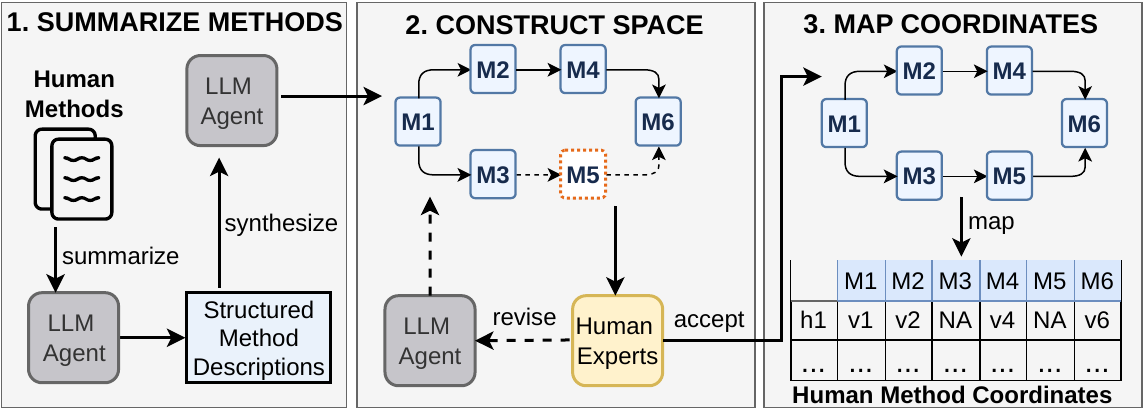}
  \caption{Human-in-the-loop construction of a task-specific algorithmic design space. LLM agents summarize human methods into structured method descriptions and use these descriptions to produce an initial version of the task-specific
  algorithmic design space. Human experts review the proposed modules, values, and dependencies, and LLM agents revise the design space based on their feedback until the revised
  design space is accepted. Each human method is then mapped to an algorithmic coordinate in the accepted design space.}
  \label{fig:ADS}
\end{figure*}
\section{Analysis Methodology}
\label{sec:analysis-methodology}

This section presents the methodology for analyzing methods designed by LLM agents in terms of both task performance and algorithmic design.
As illustrated in Figure~\ref{fig:figure1}, an agent solves an AI task by developing and executing code that produces submissions.
The submissions are evaluated using a task-specific performance metric, while the code is used to map the corresponding method into a task-specific algorithmic design space  constructed from human reference methods.

\subsection{Algorithmic Representation}
Performance metrics show how well an agent-designed method performs, but reveal little about its underlying design choices or how they compare with those of human-designed methods. 
Direct comparisons based on textual similarity are unsuitable, since equivalent algorithms may be implemented differently, while methods with distinct core designs may share substantial surface-level similarities.
A meaningful comparison therefore requires a common representation that abstracts away implementation-specific details while preserving core algorithmic choices. 
To this end, this paper represents human-designed and agent-designed methods within task-specific algorithmic design spaces, making these heterogeneous methods directly comparable.

For a task $t$, its algorithmic design space is represented as:
\begin{equation}
  \mathcal{S}_t =
  \left(
  G_t,
  \{\mathcal{O}_{t,v}\}_{v\in V_t}
  \right),
\end{equation}
where $G_t=(V_t,E_t)$ is a directed acyclic graph describing the main data flow of methods for the task.
Each node $v\in V_t$ represents an algorithmic module, each edge represents a dependency between modules, and $\mathcal{O}_{t,v}$ contains the recorded design choices for module $v$.

A method $m$ is represented by an algorithmic coordinate
\begin{equation}
  \mathbf{x}_t(m)
  =
  \left(x_{t,v}(m)\right)_{v\in V_t},
\end{equation}
where $x_{t,v}(m)\in\mathcal{O}_{t,v}$ specifies its choice at module $v$.
The coordinate captures algorithm-level choices while abstracting away implementation details such as hyperparameters and minor code-level differences.

For two methods represented in the same design space, their algorithmic distance is defined as the Hamming distance between their coordinates:
\begin{equation}
      d_t(m_1,m_2)
      =
      \sum_{v\in V_t}
      \mathbb{I}
      \left[
      x_{t,v}(m_1)\neq x_{t,v}(m_2)
      \right].
\end{equation}
This distance counts the number of modules in which the two methods make different algorithmic choices.

Let $\mathcal{H}_t$ denote the collected human reference methods for task $t$.
To position an agent-designed method $a$ relative to existing human designs, this paper defines its distance to the human reference set as
\begin{equation}
d_t(a,\mathcal{H}_t)
=
\min_{h\in\mathcal{H}_t}
d_t(a,h).
\end{equation}
A distance of zero indicates that the agent-designed method shares an algorithmic coordinate with at least one collected human method, whereas larger values indicate differences across more modules.
Because each term in the distance corresponds to a specific module, the comparison reveals not only how many algorithmic choices differ, but also where these differences occur.
\begin{table*}[htbp]
\centering
\small
\begin{tabular}{llllcc}
\toprule
Task & Domain & Dataset & Metric & SOTA year & References \\
\midrule
Image classification & Vision & CUB-200-2011~\cite{CUB200} & Accuracy $\uparrow$ & 2023 & 44 \\
Emotion classification & NLP & GoEmotions~\cite{GoEmotions} & Macro-F1 $\uparrow$ & 2026 & 24 \\
Node classification & Graph & ogbn-arxiv~\cite{OGB-Benchmark} & Accuracy $\uparrow$ & 2023 & 81 \\
Link prediction & Graph & ogbl-ppa~\cite{OGB-Benchmark} & Hits@100 $\uparrow$ & 2025 & 27 \\
Time-series forecasting & Time series & ETTh1~\cite{Informer} & MSE $\downarrow$ & 2026 & 76 \\
Time-series forecasting & Time series & Weather~\cite{Autoformer} & MSE $\downarrow$ & 2025 & 75 \\
\bottomrule
\end{tabular}
\caption{Overview of the task collection.}
\label{tab:tasks}
\end{table*}

\subsection{Design Space Construction}
This subsection describes how task-specific algorithmic design spaces are constructed from collected human reference methods and subsequently used to represent both human- and agent-designed methods.

Figure~\ref{fig:ADS} illustrates this human-in-the-loop construction process.
LLM agents first convert each human reference method into a structured description of its algorithmic design, using the corresponding code as the primary source of evidence.
Next, the LLM agents compare these descriptions across methods and group operations that serve the same role into shared modules.
Within each module, alternative operations are recorded as candidate values, while dependencies between modules are inferred from the data flow of the corresponding methods.

Human experts with relevant domain knowledge then review the extracted data flows and the proposed organization of the design space from a global perspective.
They adjudicate unresolved structural choices and provide corrective feedback when the framework does not adequately capture the collected human methods.
The LLM agents revise the design space based on these decisions, and the revised version is reviewed again until it is accepted by the human experts.

Once the design space is accepted, LLM agents assign each human reference method an algorithmic coordinate and map each agent-designed method into the same space using the finalized module definitions and dependencies.
Human experts then inspect a sample of the mappings for agent-designed methods to identify systematic assignment errors and apply the corresponding corrections consistently across the affected mappings, using the executed code as the primary evidence.
When an agent-designed method introduces a choice not observed in the human references but compatible with an existing module, that choice is added as a new value.
If the method cannot be expressed by the existing module structure, it is classified as \emph{out of space} (OOS).

The resulting spaces are not intended to reproduce every implementation detail or provide fully precise descriptions of individual algorithms.
Instead, they are designed to capture the core algorithm-level choices relevant to a shared and auditable comparison, allowing differences between human- and agent-designed methods to be traced to specific modules.

\subsection{Task Collection}
\label{subsec:task}
This subsection introduces the task collection used to study how LLM agents design methods for complex AI tasks.
It presents the task selection principles and summarizes the resulting suite.

\textbf{Task Design Principles.}
The task collection follows four design principles.
First, the collection should span representative AI tasks across multiple domains and modalities.
Second, each task should be open-ended, admitting multiple plausible algorithmic approaches rather than prescribing a single solution path.
Third, each task should leave meaningful room for improvement over existing methods.
Operationally, only tasks with a human SOTA method published between 2022 and 2026 are considered.
Fourth, each task should have sufficient published human methods to support a manually curated human leaderboard, with corresponding papers and available code collected for its entries.
The curated leaderboard provides a human performance baseline, while the collected references support the construction of task-specific algorithmic design spaces.

\textbf{Task Suite.}
The resulting suite contains six tasks spanning computer vision, natural language processing, graph learning, and time-series forecasting.
Each task is packaged as a self-contained project containing a task description, prepared data, an example submission, an evaluation script, a human leaderboard, and corresponding reference papers and
available code.
Across the six tasks, the collection contains 327 human reference entries.
Table~\ref{tab:tasks} summarizes the task suite, where SOTA year denotes the publication year of the top-ranked method on the corresponding human leaderboard~\cite{FGVC,LexEnhance,SimTEG,GraphGPT,GraFT,Towards}, and References denotes the number of collected reference entries.
\section{Experiments}
\label{sec:experiments}

\begin{table*}[htbp]
\centering
\small
\begin{tabular}{llccccccc}
\toprule
\multirow{2}{*}{Agent} & \multirow{2}{*}{Ref.} & \multicolumn{6}{c}{Performance on Each Task} & \multicolumn{1}{c}{Overall} \\
\cmidrule(lr){3-8}\cmidrule(l){9-9} & & CUB200 $\uparrow$ & GoEmotions $\uparrow$ & ogbn-arxiv $\uparrow$ & ogbl-ppa $\uparrow$ & ETTh1 $\downarrow$ & Weather $\downarrow$ & $\overline{\mathrm{Rank}}$ \\
\midrule
\textbf{Human SOTA} & -- & 0.9310 & 0.5518 & \textbf{0.7803} & \textbf{0.7655} & \textbf{0.3748} & 0.2170 & \textbf{2.67} \\
\midrule
CC-Opus & w/o & \textbf{\underline{0.9355}} & \underline{0.5553} & 0.7458 & 0.6488 & 0.5047 & 0.2518 & 5.17 \\
CC-Opus & w/ & 0.9241 & \textbf{\underline{0.5635}} & 0.7709 & 0.6288 & 0.5266 & 0.2436 & 5.00 \\
Codex & w/o & 0.8234 & 0.5304 & 0.7431 & 0.1824 & 0.5281 & 0.2296 & 10.00 \\
Codex & w/ & 0.9089 & 0.5411 & 0.7413 & 0.5049 & 0.5437 & 0.2399 & 9.83 \\
Gemini CLI & w/o & 0.9187 & \underline{0.5561} & 0.7307 & 0.5530 & 0.5137 & 0.2326 & 7.00 \\
Gemini CLI & w/ & 0.9268 & \underline{0.5629} & 0.7686 & 0.6208 & 0.5297 & 0.2267 & 4.83 \\
CC-DeepSeek & w/o & 0.9172 & 0.5510 & 0.7308 & 0.5665 & 0.5154 & 0.2405 & 8.33 \\
CC-DeepSeek & w/ & 0.8963 & \underline{0.5633} & 0.7656 & 0.4947 & 0.5068 & 0.2321 & 6.33 \\
CC-GLM & w/o & 0.9299 & \underline{0.5596} & 0.7467 & 0.5008 & 0.5119 & 0.2227 & 4.92 \\
CC-GLM & w/ & 0.9299 & \underline{0.5557} & 0.7657 & 0.5124 & 0.5271 & \textbf{\underline{0.2070}} & 4.92 \\
MLEvolve & w/o & 0.9237 & \underline{0.5545} & 0.6952 & 0.1962 & 0.7021 & 0.2347 & 10.00 \\
MLEvolve & w/ & 0.9077 & 0.5417 & 0.7148 & 0.0966 & 0.6829 & 0.2797 & 12.00 \\
\bottomrule
\end{tabular}
\caption{Best performance achieved within the evaluation budget for each agent, task, and reference condition. The Ref. column indicates whether prepared reference papers and corresponding code are provided (w/) or withheld (w/o).
CC-Opus, CC-DeepSeek, and CC-GLM denote Claude Code using Claude Opus 4.8, DeepSeek 4 Pro, and GLM 5.2, respectively.
Codex and MLEvolve use GPT 5.5, while Gemini CLI uses Gemini Flash 3.5.
The Human SOTA row reports the best collected human result for each task.
Upward and downward arrows indicate whether higher or lower values are better.
$\overline{\mathrm{Rank}}$ is the mean task-wise rank across Human SOTA and the twelve agent configurations.
\textbf{Bold} values indicate the best result in each column, while \underline{underlined} agent results meet or exceed Human SOTA.}
\label{tab:performance-overview}
\end{table*}

This section empirically evaluates widely used LLM agents as designers of methods for complex AI tasks.
It asks three questions: how closely agent-designed methods approach the human frontier, how their algorithmic designs differ from existing human methods, and what characterizes the methods that reach that
frontier.
After describing the experimental setup, the analysis addresses these questions through task-level performance, algorithmic distance, and a focused examination of frontier-reaching methods.
Finally, as a complementary analysis, the section examines whether prepared human references improve agent performance and whether agents actively seek and use external knowledge.

\subsection{Experimental Setup}
\label{subsec:experimental-setup}
The evaluation comprises 72 experimental configurations, covering all \(6 \times 6 \times 2\) combinations of (agent, task, reference condition).
In each configuration, an agent receives the task description, prepared data, an example submission, and an evaluation interface, develops and executes an end-to-end method, and submits its predictions for evaluation.
The evaluator returns only a scalar performance score without exposing individual test labels or targets.
Each evaluated method is recorded as a solution, with a complete snapshot of its code, runnable script, submission file, method summary, and execution logs.
Beyond task-specific performance, these methods are mapped into the corresponding algorithmic design spaces, while available agent traces are used to examine how agents conduct research.
The evaluated agents and remaining experimental settings are described below.

\textbf{Agents.}
Three Claude Code agents~\cite{ClaudeCode} use Claude Opus 4.8~\cite{Opus-4.8}, DeepSeek 4 Pro~\cite{DeepSeekV4}, and GLM 5.2~\cite{GLM-5.2};
Codex~\cite{Codex} uses GPT 5.5~\cite{GPT-5.5};
Gemini CLI~\cite{GeminiCLI} uses Gemini Flash 3.5~\cite{Gemini-Flash-3.5};
and MLEvolve~\cite{MLEvolve} uses GPT 5.5.

\textbf{Tasks.}
The six tasks are those introduced in Section~\ref{subsec:task}.
For brevity, CUB-200-2011 is referred to as CUB200.

\textbf{Reference Conditions.}
Each (agent, task) pair is evaluated both with and without prepared reference papers and corresponding code.
Internet access and all other available tools remain enabled under both conditions.

\textbf{Execution Budget.}
Each experimental configuration is allocated up to 24 hours on a single NVIDIA V100 GPU.
Within this budget, the agent may iteratively produce and evaluate up to 10 successive solutions.

\begin{figure*}[t]
  \centering
  \begin{subfigure}[t]{0.48\textwidth}
    \centering
    \includegraphics[width=\linewidth]{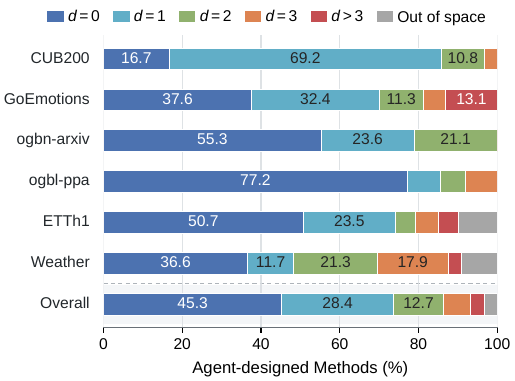}
    \caption{Algorithmic distance distributions by task.}
    \label{fig:distance-distribution-task}
  \end{subfigure}
  \hfill
\begin{subfigure}[t]{0.49383\textwidth}
    \centering
    \includegraphics[width=\linewidth]{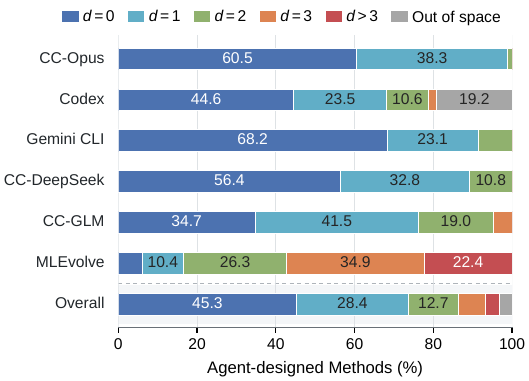}
    \caption{Algorithmic distance distributions by agent.}
    \label{fig:distance-distribution-agent}
  \end{subfigure}  
  \caption{Distributions of the minimum algorithmic distance from agent-designed methods to collected human methods, group separately by agent and task.}
  \label{fig:distance-distributions}
\end{figure*}

\subsection{Performance}
\label{subsec:performance-analysis}

This subsection examines how closely agent-designed methods approach the human performance frontier across agents and tasks.
Table~\ref{tab:performance-overview} reports the best performance achieved in each experimental configuration and summarizes cross-task performance using average rank.
Two main findings emerge from this comparison.

\textbf{Current agents occasionally reach the human frontier, but such success does not generalize reliably.}
Among the 72 experimental configurations, 10 reach or surpass Human SOTA, but these successes occur on only 3 of the 6 tasks.
Eight are concentrated on GoEmotions, while CUB200 and Weather each account for only one.
Moreover, no agent under either reference condition reaches Human SOTA on more than two tasks.
Human SOTA retains the best average rank at 2.67, leaving a clear gap to the strongest individual agent configuration, Gemini CLI with prepared references, at 4.83.
When each agent's ranks are averaged across the two reference conditions, the overall ordering is CC-GLM, CC-Opus, Gemini CLI, CC-DeepSeek, Codex, and MLEvolve.

\textbf{The gap to the human frontier varies substantially across tasks.}
Agents perform most strongly on GoEmotions, where 8 of the 12 configurations surpass Human SOTA.
CUB200 and Weather each exhibit only one frontier-reaching result, while all configurations on ogbn-arxiv remain below but relatively close to Human SOTA.
The largest gaps occur on ETTh1 and ogbl-ppa, where the best agent results still have 34.7\% higher MSE and 15.2\% lower Hits@100 than Human SOTA, respectively.
These results show that the ability of current agents to reach the human frontier remains strongly task-dependent.

\subsection{Algorithmic Exploration}
\label{subsec:behavioral-analysis}

This subsection examines agent-designed methods from two complementary perspectives: their positions relative to existing human methods and the breadth of their exploration within the task-specific algorithmic design spaces.

\textbf{Agent-designed methods generally remain close to existing human methods.}
Figures~\ref{fig:distance-distribution-task} and~\ref{fig:distance-distribution-agent} report the algorithmic distance distributions by task and agent, respectively.
Across all agents and tasks, 45.3\% of the methods share an algorithmic coordinate with at least one collected human method, while another 28.4\% differ from the nearest human method in exactly one module.
Thus, 73.7\% lie within \(d\leq1\) of a human method.
This proximity is observed across most tasks: five of the six tasks have at least 70\% of their methods within \(d\leq1\), with Weather as the main exception at 48.3\%.
The distributions also vary across agents.
For the five agents other than MLEvolve, the proportion of methods within \(d\leq1\) ranges from 68.1\% to 98.8\%.
MLEvolve is a clear exception, with only 16.4\% within \(d\leq1\) and 57.2\% differing from the nearest human method in at least three modules.

\textbf{Most agent-designed methods are recombinations or local variations of human designs.}
Among methods represented within the design spaces, 95.3\% of module-level choices use values already observed in collected human methods, while only 4.7\% require previously unobserved values.
Moreover, only 3.2\% of the methods fall outside the constructed design spaces.
Together with the distance distributions, these results indicate that agent-designed methods primarily reproduce or recombine established human choices.

\begin{figure}[t]
\centering
\includegraphics[width=\columnwidth]{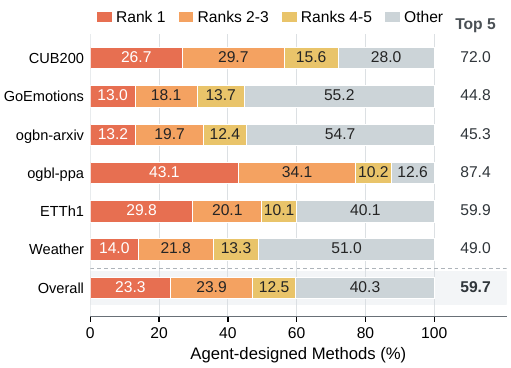}
\caption{Concentration of agent-designed methods on the most frequently visited algorithmic coordinates within each task.
Rank 1 denotes the most frequent coordinate, while Ranks 2--3 and Ranks 4--5 denote the corresponding groups of coordinates.
Other includes less frequent coordinates and methods outside the constructed design spaces.
Overall reports the mean distribution across the six tasks.}
\label{fig:coordinate-concentration}
\end{figure}

\textbf{Agents repeatedly explore a limited set of algorithmic coordinates.}
Figure~\ref{fig:coordinate-concentration} reports the proportions of agent-designed methods assigned to the five most frequent algorithmic coordinates in each task.
Although each experimental configuration can produce up to ten successive solutions, these solutions occupy a median of only 3.5 distinct coordinates.
The five most frequent coordinates account for more than half of all agent-designed methods in four of the six tasks and 59.7\% on average.
These results indicate that agents repeatedly return to a small set of algorithmic designs rather than broadly covering the constructed design spaces.

\subsection{Frontier-Reaching Methods}
\label{subsec:performance-behavior}

This subsection analyzes how high-performing agent-designed methods relate algorithmically to existing human methods.

\begin{figure}[t]
  \centering  \includegraphics[width=\columnwidth]{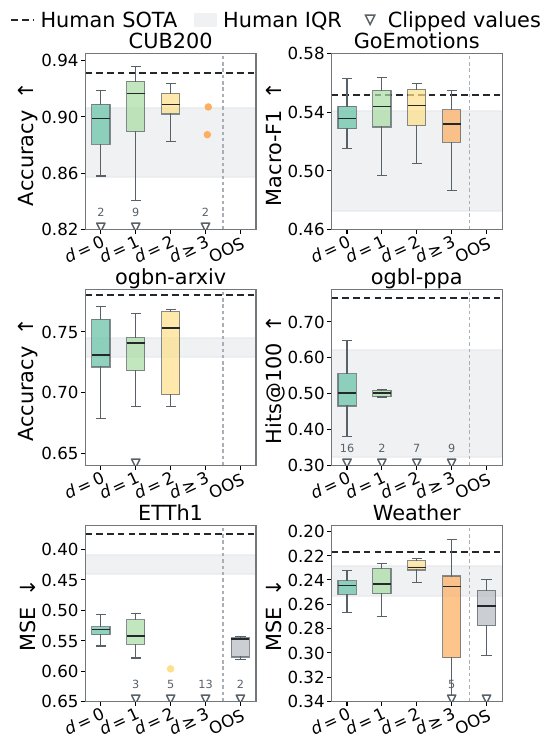}  
  \caption{Task performance across algorithmic distance categories. Boxes show the performance distributions for categories containing at least eight solutions, while individual points are shown for smaller categories. Dashed lines indicate Human SOTA, while the regions labeled Human IQR span the 25th to 75th percentiles of the collected human performance results. OOS denotes solutions outside the constructed design spaces. Open triangles indicate results beyond the displayed vertical range, with numbers reporting their counts when greater than one.}
  \label{fig:performance-behavior}
\end{figure}

\textbf{High-performing methods generally remain close to existing human designs.}
Figure~\ref{fig:performance-behavior} compares the performance of agent-designed methods across algorithmic distance categories.
None of the 23 solutions outside the constructed design spaces ranks in the top quartile for its task.
Moreover, in five of the six tasks, the best-performing agent-designed method lies within \(d\leq1\).
Weather is the only exception, with its best-performing method located at \(d=3\).

\textbf{Methods reaching Human SOTA generally recombine human-observed choices and rely on ensemble prediction.}
Among the 10 configurations that reach or surpass Human SOTA, 8 of them produce a SOTA-reaching method at \(d\leq2\), and none produces one outside the constructed design spaces.
Moreover, 88.0\% of the module-level choices in their best methods use values already observed in collected human methods.
Notably, the best method in each of the 10 SOTA-reaching configurations uses ensemble prediction.

\subsection{Use of Human Knowledge}

This subsection examines whether providing agents with prepared human references helps them design methods for these tasks and whether agents actively seek and use existing human knowledge.

\textbf{Prepared references do not consistently improve performance.}
Table~\ref{tab:performance-overview} provides 36 matched comparisons between experimental configurations that differ only in whether prepared reference papers and corresponding code are provided.
Providing references improves performance in 20 comparisons, reduces it in 15, and leaves it unchanged in 1.
Moreover, among the 10 configurations that reach or surpass Human SOTA, 5 use prepared references and the other 5 do not.
Thus, under the experimental setup of this study, providing prepared references does not yield a clear improvement in task performance or increase the number of configurations that reach or surpass Human SOTA.

\textbf{Agents make limited use of external knowledge.}
Among the agents studied, MLEvolve is the only one without the ability to actively access the Internet; all other agents can search for relevant online resources, and Internet use is unrestricted in the experiments.
Nevertheless, only 2 of the 60 configurations involving these five agents conduct task-specific online searches for existing methods.
Even when references are provided, agents access only 5.4\% of them and consult a top-3 reference for the corresponding task in only 8 of the 36 configurations.
These results suggest that agents rely primarily on knowledge learned during LLM pretraining rather than systematically surveying existing human methods.
\section{Discussion}
\label{sec:discussion}

Most methods produced by current agents can be understood as recombinations of algorithmic choices observed in human methods.
Methods that reach or surpass Human SOTA also remain predominantly close to existing human designs, although Weather shows that larger departures can occasionally succeed.
These successes, however, remain task-dependent and coexist with limited exploration of the algorithmic design spaces and limited use of external knowledge.
Taken together, these findings indicate that current agents can occasionally produce strong methods but cannot yet do so reliably across tasks.

These findings raise a broader question: how can agents turn occasional successes into reliable progress?
From the perspective of algorithmic design spaces, such progress may follow three broad routes, depending on whether it preserves an existing coordinate, moves to another coordinate, or expands beyond the existing space:
(1) \textbf{coordinate-preserving optimization}, which adjusts parameters or implementation details while keeping the coordinate fixed;
(2) \textbf{within-space recombination}, which moves to a new coordinate primarily by combining established module choices;
(3) \textbf{space expansion}, which seeks designs that substantially extend or fall outside the constructed space.

Coordinate-preserving optimization is the most accessible, but its performance ceiling remains bounded by the selected coordinate.
Space expansion offers the highest theoretical ceiling because it is not constrained by the existing design space, but it is also the most difficult route to pursue reliably.
Within-space recombination offers a trade-off, with a higher performance ceiling than coordinate-preserving optimization and greater feasibility than space expansion.
The experimental results in this paper also demonstrate the feasibility of this route, as most methods that reach or surpass Human SOTA appear to follow it.
\section{Conclusion}
\label{sec:conclusion}

This paper investigates how well current LLM agents can design methods for complex AI tasks and how their algorithmic designs relate to existing human methods.
To support this investigation, an analytical perspective based on algorithmic design spaces is introduced, through which human- and agent-designed methods are mapped into shared task-specific spaces for module-level comparison.
Widely used LLM agents are evaluated on a suite of representative, open-ended AI tasks spanning multiple modalities.
The results show that current agents can occasionally reach or surpass Human SOTA, but such success remains task-dependent and unreliable, while most agent-designed methods primarily recombine algorithmic choices observed in human methods.
Finally, this paper outlines three possible routes toward more reliable progress: coordinate-preserving optimization, within-space recombination, and space expansion.

{
    \small
    \bibliographystyle{ieeenat_fullname}
    \bibliography{main}
}

\end{document}